\documentclass[11pt,table]{article}
\usepackage[final]{acl}

\usepackage{xcolor}
\usepackage{times}
\usepackage{latexsym}
\usepackage[utf8]{inputenc}
\usepackage{microtype}
\usepackage{inconsolata}
\usepackage{graphicx}
\usepackage{amssymb}
\usepackage{amsmath}
\usepackage{amsthm}
\usepackage{subcaption}
\usepackage{listings}
\usepackage{float}
\usepackage[T1]{fontenc}
\usepackage[utf8]{inputenc}
\usepackage{textcomp}
\title{Test of Time: Rethinking Temporal Signal of Benchmark Contamination}

\author{
Terry Jingchen Zhang\textsuperscript{1,2*}
\quad 
Gopal Dev\textsuperscript{1,3*}
\quad 
Ning Wang\textsuperscript{2}
\quad 
Max Obreiter\textsuperscript{2}
\\\textbf{
Punya Syon Pandey\textsuperscript{1}
\quad 
Keenan Samway\textsuperscript{1,3}
\quad 
Wenyuan Jiang\textsuperscript{2}
\quad 
Yinya Huang\textsuperscript{2}
}
\\\textbf{
Bernhard Sch\"olkopf\textsuperscript{3,4}
\quad 
Mrinmaya Sachan\textsuperscript{2}
\quad 
Zhijing Jin\textsuperscript{1,3}
}
\vspace{0.3em}
\\
\textsuperscript{1}Jinesis Lab, University of Toronto \& Vector Institute\\
\textsuperscript{2}ETH Z\"urich \& ETH AI Center\\
\textsuperscript{3}Max Planck Institute for Intelligent Systems, T\"ubingen, Germany\\
\textsuperscript{4}ELLIS Institute T\"ubingen
\\
\vspace{-2em}
}

\usepackage{hyperref}       
\usepackage{cleveref}
\usepackage{url}            
\usepackage{booktabs}       
\usepackage{graphicx}
\usepackage{tabularx}
\usepackage{enumitem}
\usepackage{bm}

\usepackage[textsize=tiny]{todonotes}
\newlist{dialogue}{description}{1}
\setlist[dialogue]{
    labelwidth=1.5cm,
    labelindent=0cm,
    leftmargin=1.8cm,
    labelsep=0.3cm,
    align=left,
    noitemsep,
    topsep=0pt
}

\definecolor{colInter}{HTML}{1F77B4}    
\definecolor{colCorr}{HTML}{2CA02C}   
\definecolor{colErrDesc}{HTML}{FF7F0E}  
\definecolor{colInst}{HTML}{D62728}     
\definecolor{colPlaus}{HTML}{9467BD}  
\definecolor{colCur}{HTML}{17BECF}    

\begin{document}

\maketitle

\begingroup
\renewcommand\thefootnote{}
\footnotetext{*Equal contribution.
Our codes are available at \url{https://github.com/AI4Collaboration/Test-of-Time}.
}
\endgroup

\begin{abstract}
Post-cutoff performance decay of LLMs has been widely interpreted as a temporal signal for benchmark contamination, where public information released before the training cutoff may have been included into training corpora and inflated model performance by memorization.
We critically examine this view and demonstrate that this temporal signal is highly sensitive to how benchmark questions are constructed, even if the underlying source material remains invariant.
Specifically, we show that LLM-transformed questions can produce remarkably different temporal patterns compared to fill-in-the-blank (cloze) questions directly retrieved from the very same documents.
We validate this effect on prior benchmarks that report clear post-cutoff decay (LiveCodeBench), and show that a simple LLM-driven transformation of the same problems can effectively remove the temporal pattern.
We further provide a mechanistic understanding of this phenomenon using influence function analysis.
Overall, our results suggest that post-cutoff performance decay is a sensitive contamination signal, motivating more robust contamination probes for reliable LLM evaluation.
\end{abstract}

\section{Introduction}
As frontier large language models are trained on increasingly larger datasets, publicly available benchmarks may be included into massive web-scale training corpora~\cite{BenchmarkDataContamination,ContaminationSystematic}. This leads to the problem of \textbf{benchmark contamination}, where models may answer evaluation questions by memorizing leaked items rather than reasoning. As a result, contamination poses a fundamental threat to the reliability of evaluation: it can conflate memorization with genuine capability~\cite{ContaminationSpectrum,GeneralizationOrMemorization}.

\begin{figure}[!t]
    \centering
    \includegraphics[width=1\linewidth]{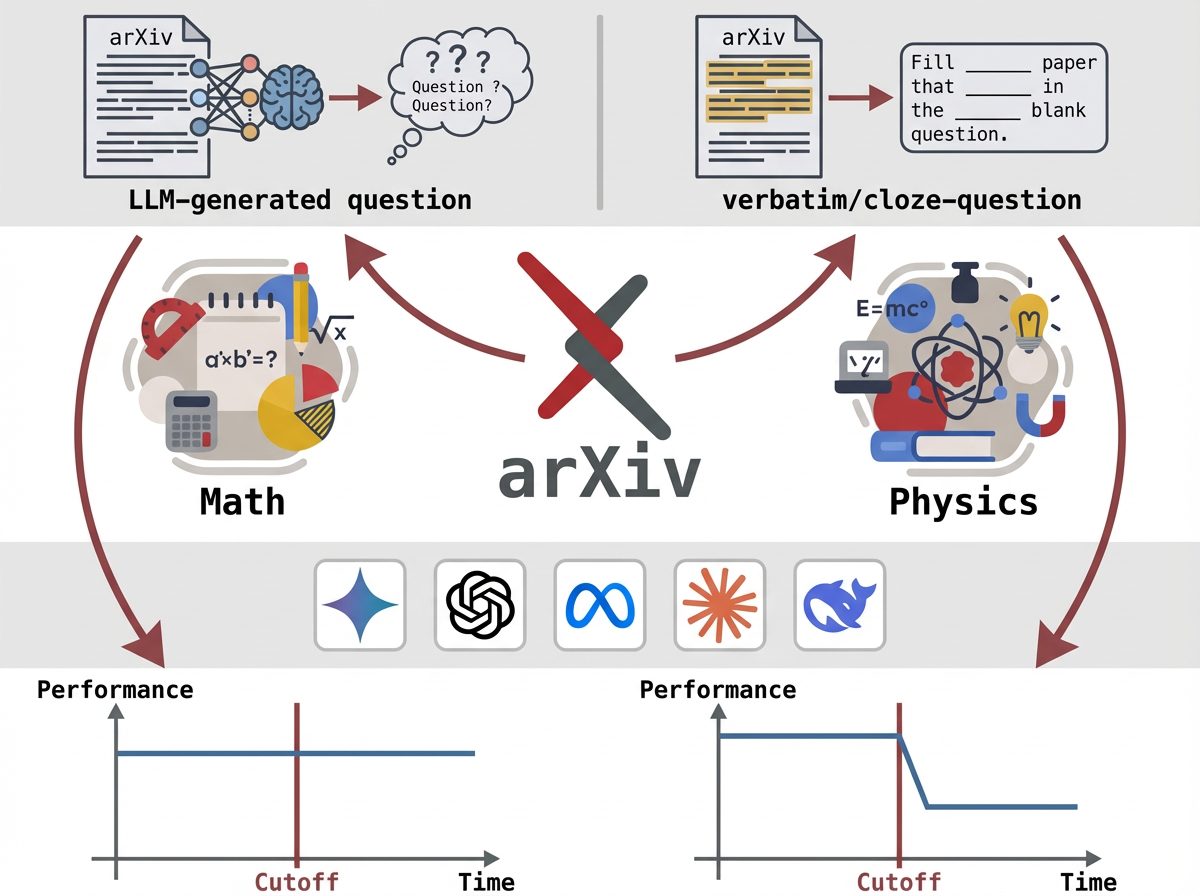}
    \caption{Overview of the temporal analysis framework: based on the same underlying source material (arXiv papers in math and physics), we compare the temporal pattern of model performance in 2 types of QAs: verbatim/cloze questions directly retrieved from public sources may demonstrate post-cutoff performance decay, whereas questions generated/transformed by LLMs do not.}
    \label{fig:framework}
\end{figure}

To detect such contamination, a growing line of work generates variants of existing benchmarks by way of rephrasing~\cite{DBLP:journals/tacl/GolchinS25, xu2025reimagine}, expert annotation~\cite{GSM1K,MATH-Perturb}, or LLM-based synthesis of new questions~\cite{AlpacaVicuna,DVD}.
Most previous work in this direction consistently reported performance degradation on variants comparing to the original benchmark, a phenomenon widely interpreted as evidence to the brittleness of LLM reasoning and potential contamination~\cite{djiré2025memorizationinterpolationdetecting}.

Another widely used approach is \emph{temporal analysis}, which compares performance on questions released before vs. after a model's training cutoff date~\cite{TaskContam,TimeTravelContamination}.
It's considered as a simple yet intuitive contamination probe: if models perform worse on questions released after their cutoff dates (which we term as \textbf{post-cutoff performance decay}), this gap is frequently interpreted as evidence that pre-cutoff questions (or close variants) leaked into training data.
Previous work on this line has reported such post-cutoff performance decay on questions directly retrieved from public websites across multiple domains, including math, coding and reasoning at large~\cite{LiveCodeBench,LiveBench,CutoffAndBeyond}.

However, a recent study~\cite{zhang2025realmath} reported a counter-intuitive \emph{lack} of post-cutoff decay on LLM-generated questions from mathematics arXiv papers, which is commonly present in web-scale training corpora used for frontier models. This discrepancy raises an intriguing question: why can temporal analysis show clear decay on verbatim benchmarks like LiveCodeBench~\cite{LiveCodeBench}, yet fail to show decay on LLM-synthesized questions from time-stamped sources?

As illustrated in Figure~\ref{fig:framework}, we first investigated this counter-intuitive observation by validating the lack of post-cutoff performance decay at scale: specifically, we extended temporal analysis to 1,643 LLM-generated questions from 20,277 arXiv papers spanning 26 months across 8 models and 2 domains (mathematics and physics), confirming that the lack of post-cutoff decay is a consistent pattern across different model families and datasets.
We hypothesize that the key factor behind this phenomenon may be the LLM generation process that transformed the base material (the same arXiv papers) enough such that models can no longer recall them from training corpora.

We put this hypothesis to the test by constructing cloze (fill-in-the-blank) questions using the same set of papers and find that post-cutoff performance decay appears even though both cloze and LLM-generated questions are based on the same source papers.
Furthermore, we validated this hypothesis beyond arXiv papers by showing that LLM generation could also effectively remove post-cutoff decay on LiveCodeBench~\cite{LiveCodeBench} (which previously reported clear post-cutoff performance decay) and a FreshQA-style dataset~\cite{vu-etal-2024-freshllms} based on Wikipedia without changing the underlying solution.

Finally, we use influence function analysis on provably contaminated training documents to provide a mechanistic understanding of our observation: when given verbatim/cloze questions simply copy-pasted from arXiv papers, models were able to identify source documents among the most influential training documents, yet LLM-generated QA based on the same set of papers were much harder to identify.

Taken together, our findings call for a careful re-examination of post-cutoff performance decay as a temporal signal for benchmark contamination, as it can be highly sensitive to different benchmark construction methodologies, among other potential confounders.
We further argue that future studies should explore more robust contamination probes under various benchmark construction methods.

\begin{table*}[h]
\caption{Examples of synthesized QA questions and corresponding CLOZE questions from the same arXiv papers.}
\centering
\renewcommand{\arraystretch}{1.4}
\definecolor{mathcolor}{RGB}{225,242,245} 
\definecolor{physcolor}{RGB}{225,245,238} 
\definecolor{papercolor}{RGB}{232,245,250} 
\definecolor{clozecolor}{RGB}{200,235,220} 
\setlength{\tabcolsep}{5pt}
\scalebox{0.88}{
\begin{tabular}{|p{17.6cm}|}
\hline
\multicolumn{1}{|c|}{\cellcolor{papercolor}\textbf{Math} --- \href{http://arxiv.org/abs/2406.19979v2}{arXiv:2406.19979v2} - On quantitative convergence for stochastic processes} \\
\hline
\textbf{Synthesized QA:} Given parameters $\varepsilon>0$, $K>0$, and a function $g:\mathbb{N}\to\mathbb{N}$, what explicit bound $N$ in terms of $\varepsilon$, $K$, and $g$ guarantees that for any monotone sequence $(x_n)$ in $[-K,K]$ there exists some $n\le N$ such that
$|x_i - x_j|<\varepsilon \quad\text{for all }n\le i\le j\le n + g(n)?$ \\
\hline
\multicolumn{1}{|c|}{\cellcolor{clozecolor}CLOZE Question (from abstract)} \\
\hline
We develop a general framework for extracting highly uniform bounds on local stability for \colorbox{yellow!30}{[blank1]} processes in terms of information on fluctuations or crossings. This includes a large class of \colorbox{yellow!30}{[blank2]}: As a corollary of our main abstract result, we obtain a quantitative version of Doob's convergence theorem for \colorbox{yellow!30}{[blank3]}-sub- and supermartingales, but more importantly, demonstrate that our framework readily extends to more complex stochastic processes such as \colorbox{yellow!30}{[blank4]}, thus paving the way for future applications in stochastic optimization. Fundamental to our approach is the use of ideas from logic, particularly a careful analysis of the \colorbox{yellow!30}{[blank5]} structure of probabilistic statements and the introduction of a number of abstract notions that represent stochastic convergence in a quantitative manner. In this sense, our work falls under the 'proof mining' program, and indeed, our quantitative results provide new examples of the phenomenon, recently made precise by the first author and Pischke, that many proofs in probability theory are proof-theoretically tame, and amenable to the extraction of quantitative data that is both of low complexity and independent of the underlying probability space.\\
\hline
\multicolumn{1}{|c|}{\cellcolor{papercolor}\textbf{Physics} --- \href{http://arxiv.org/abs/2407.02415v2}{arXiv:2407.02415v2} - The Symplectic Schur Process} \\
\hline
\textbf{Synthesized QA:} Let $i(\tau)=\left\lfloor\tfrac{9n}{8}+\sqrt{\tfrac{27n}{64}}\,\tau\right\rfloor$ and $u(\alpha)=-n+\left\lfloor\bigl(\tfrac{n}{12}\bigr)^{1/4}\alpha\right\rfloor$. For a fixed $k\in\mathbb{Z}_{\ge1}$ and real parameters $\tau_1,\dots,\tau_k$ and $\alpha_1,\dots,\alpha_k$, define $i_\ell=i(\tau_\ell)$ and $u_\ell=u(\alpha_\ell)$. What is the limit as $n\to\infty$ of
$$\det_{1\le\ell,\ell'\le k}\Bigl[\bigl(\tfrac{n}{12}\bigr)^{1/4}\bigl(\delta_{i_\ell,i_{\ell'}}\,\delta_{u_\ell,u_{\ell'}}-K^{\mathrm{SSP}}(i_\ell,u_\ell;i_{\ell'},u_{\ell'})\bigr)\Bigr]?$$ \\
\hline
\multicolumn{1}{|c|}{\cellcolor{clozecolor}CLOZE Question (from abstract)} \\
\hline
We define a measure on tuples of partitions, called the \colorbox{yellow!30}{[blank1]} Schur process, that should be regarded as the right analogue of the Schur process of \colorbox{yellow!30}{[blank2]} for the Cartan type C. The weights of our measure include factors that are universal symplectic characters, as well as a novel family of ``Down-Up Schur functions'' that we define and for which we prove new identities of \colorbox{yellow!30}{[blank3]}-type. Our main structural result is that the point process corresponding to the symplectic Schur process is \colorbox{yellow!30}{[blank4]} and we find an explicit correlation kernel. We also present dynamics that preserve the family of symplectic Schur processes and explore an alternative sampling scheme, based on the \colorbox{yellow!30}{[blank5]} insertion algorithm, in a special case. Finally, we study the asymptotics of the Berele insertion process and find explicit formulas for the limit shape and fluctuations near the bulk and the edge. One of the limit regimes leads to a new kernel that resembles the symmetric Pearcey kernel.\\
\hline
\end{tabular}
}
\label{tab:theorem_qa_conversion}
\end{table*}

\section{Related Work}
\paragraph{Benchmark Contamination.}
Benchmark contamination refers to overlap between a model's training data and evaluation benchmarks, which can inflate scores via memorization rather than genuine reasoning~\cite{GeneralizationOrMemorization}. This overlap can take several forms, including exact inclusion, near-duplicates, and derivative content~\cite{BenchmarkDataContamination,ContaminationSpectrum}.
Unfortunately, direct contamination audit is extremely difficult at scale~\cite{DataContaminationSurvey} because most frontier model developers do not publicly release their training data (\emph{open-data}) though some of them release their model weights (\emph{open-weight}).

\paragraph{Probe Contamination by LLM-Generated Evaluation.}
A widely used method for detecting contamination is to generate variants of existing benchmark questions and check whether model performance degrades~\cite{DBLP:journals/tacl/GolchinS25, xu2025reimagine} on the variant.
Previous work has explored simple paraphrasing using GPT-4o~\cite{djiré2025memorizationinterpolationdetecting}, perturbations using human expert annotators or stronger LLMs~\cite{RephraseBench,GSM1K,MATH-Perturb}, and even fully synthetic questions generated by LLMs~\cite{AlpacaVicuna,DVD}.
These studies consistently reported performance degradation on the generated variants, which showcased the brittleness of LLM performance on reasoning benchmarks.

\paragraph{Probe Contamination by Temporal Analysis.}
Temporal analysis compares performance on questions released before versus after a model's announced training cutoff~\cite{TaskContam,TimeTravelContamination}. It has been used in coding benchmarks including LiveCodeBench~\cite{LiveCodeBench} and LiveBench~\cite{LiveBench}, as well as in mathematics benchmarks with clearly dated problem releases~\cite{CutoffAndBeyond}. Beyond contamination audits, temporal splits are also used to study how quickly model knowledge goes stale and how retrieval or search augmentation can mitigate this issue~\cite{vu-etal-2024-freshllms}. These challenges also closely relate to hallucination, where models struggle due to incomplete or outdated knowledge~\cite{ji-etal-2024-llm, lee2024llm}, with temporal misalignment and hallucination arising from 
knowledge cutoff forming common failure modes. Across these settings, a recurring finding is that LLMs often perform worse on post-cutoff questions when the evaluation uses questions from public sources.

\section{Temporal Analysis as a Measure of Contamination Detection}
Temporal analysis is a widely used measure for detecting contamination: if models perform better on questions released before their knowledge cutoff than on those released after in the same dataset, the resulting gap (\emph{post-cutoff performance decay}) is often interpreted as evidence that pre-cutoff materials may have leaked into training data.

In this section, we first validate the lack of post-cutoff performance decay on RealMath~\cite{zhang2025realmath} at scale.
We extend their setting to a longer 26-month time range and multiple domains (mathematics and physics), and evaluate more diverse models with distinct developers \& cutoff dates to test whether the lack of post-cutoff decay is a consistent pattern.
We then use the resulting evidence to motivate our hypothesis, which we subsequently validate in Section~4.

\begin{table}[]
\centering
\caption{Summary of evaluated models and their knowledge cutoff dates. We carefully include 4 frontier model families (2 models per family with different cutoff dates) to test whether temporal patterns are consistent across architectures and time windows.}
\label{tab:cutoffs}
\begin{tabular}{lc} 
\hline
\textbf{Model} & \textbf{Knowledge Cutoff} \\
\hline
DeepSeek-R1-0528 & 2024.07 \\
DeepSeek-R1 & 2023.10 \\
\hline
OpenAI-o4-mini & 2024.06 \\
OpenAI-o3-mini & 2023.10 \\
\hline
Gemini-2.5-Flash & 2025.01 \\
Gemini-2.0-Flash & 2024.08 \\
\hline
Llama-4-Scout & 2024.08 \\
Llama-3.3-70B & 2023.12 \\
\hline
\end{tabular}
\end{table}

\subsection{Methodology}
We retrieve 20,277 arXiv papers using the arXiv API\footnote{\url{https://info.arxiv.org/help/api/index.html}} with complete metadata from May 2023 to June 2025 (26 months), which is selected to cover at least 6 months before and after the cutoff dates of all evaluated models. We retrieve papers from physics and mathematics domains to validate the findings across disciplinary boundaries. For physics, we specifically filter out papers from 5 subdomains, namely General Relativity and Quantum Cosmology, Mathematical Physics, Exactly Solvable and Integrable Systems, Computational Physics, and Fluid Dynamics. We found these subdomains have high density of useful theorems in our analysis and also enough paper corpus per month to generate adequate QA pairs, which is a strict requirement for a temporally balanced evaluation across months.

Following the question generation protocols of RealMath~\cite{zhang2025realmath}, we use o4-mini~\cite{o3_o4-mini} to generate multi-step reasoning questions from these theorems. The resulting QA pairs are further filtered to remove trivial or duplicated ones using GPT-4.1~\cite{openai2025gpt41}.
Each resulting question has a unique deterministic ground-truth answer and requires more than five intermediate reasoning steps to solve. We implement monthly quotas to ensure roughly equal question distribution across 26 months.

We manually inspect the questions to confirm the following quality criteria: (1) clear deterministic answers, (2) at least 5 clearly separate steps in the answer, (3) unambiguous problem statements, and (4) correct derivation from source material.
This process yields 1,098 papers producing suitable questions, with human quality checks removing 47 of 1,690 initial generated questions (2.8\% rejection rate), resulting in 1,643 LLM-generated questions.

\paragraph{Evaluation Setup.}
We evaluate 4 frontier model families (Table~\ref{tab:cutoffs}), where each family is represented by 2 models with different cutoff dates to test consistency across architectures and time windows.
We extend evaluation across two domains (mathematics and physics) to see whether patterns generalize across scientific disciplines.
All models are queried via OpenRouter API\footnote{OpenRouter (\url{https://openrouter.ai}) provides unified API access to a wide range of frontier models.} using default settings without any web search access.
To address potential concerns about how frontier models could be equipped with hidden retrieval-augmented generation mechanisms, we include both proprietary models (OpenAI~\cite{o3_o4-mini}, Gemini~\cite{GeminiFlash2.5}) and open-weight models (DeepSeek~\cite{deepseekai2025deepseekr1incentivizingreasoningcapability}, Llama~\cite{meta2025llama4, grattafiori2024llama3herdmodels}) deployed by third-party providers, which yielded consistent trends in general.

\begin{table*}[h!]
\centering
\caption{Pre- versus Post-cutoff monthly accuracy in the physics domain (averaged over the full time window of evaluation). Gap (pp) denotes $\text{Post} - \text{Pre}$ in percentage points; additional domain results are in Appendix~\ref{appendix:math_results}.}
\label{tab:pre_post_physics}
\begin{tabular}{lccc}
\hline
\textbf{Model Label} & \textbf{Pre-cutoff (\%)} & \textbf{Post-cutoff (\%)} & \textbf{Gap (pp)} \\
\hline
DeepSeek-R1        & 21.1 & 22.7 & +1.6 \\
DeepSeek-R1-0528   & 21.8 & 26.2 & +4.4 \\
Gemini-2.0-Flash   & 21.6 & 26.7 & +5.1 \\
Gemini-2.5-Flash   & 33.3 & 39.2 & +5.9 \\
Llama-3.3-70B      & 15.1 & 15.5 & +0.4 \\
Llama-4-Scout      & 14.3 & 16.8 & +2.5 \\
o3-mini            & 31.3 & 36.2 & +4.9 \\
o4-mini            & 36.8 & 40.5 & +3.7 \\
\hline
\end{tabular}
\end{table*}

\subsection{Results}

We followed standard grading protocols from RealMath~\cite{zhang2025realmath} except using o4-mini as a stronger judge model for evaluation. To account for statistical variance in questions per month, we normalize model performance as $\text{Accuracy}_m  = \frac{C_m}{Q_m}$, where $C_m$ is correct answers in month $m$ and $Q_m$ is total questions in month $m$. The aggregate trend across models shows no systematic post-cutoff performance decay.
Table~\ref{tab:pre_post_physics} outlines average pre- versus post-cutoff normalized monthly performance in physics (with additional results in Appendix~\ref{appendix:math_results}), with full monthly trajectories across the 26-month period provided in Appendix~\ref{appendix:monthly}. To mitigate monthly variance, we further aggregate performance across equal-duration time windows. Specifically, we compare the aggregate performance of $n$ months before cutoff ($nB$) with $n$ months after cutoff ($nA$) for $n \in \{2, 3, 4, 5\}$ in Appendix~\ref{appendix:aggregate_results}. Across all aggregation windows, we observe a consistent absence of post-cutoff performance decay. We further validate the statistical significance of our findings in the following paragraphs.

\paragraph{Statistical Analysis.}
We define our unit of analysis as model-aggregated performance across temporal windows, where each observation represents the mean performance difference (post-cutoff minus pre-cutoff accuracy) for one model. This yields 16 independent observations (8 models × 2 domains). While monthly observations may exhibit temporal correlation, our aggregation approach and use of multiple model families with different cutoff dates help mitigate this concern. The mean performance change across all observations was $+2.19$ percentage points (95\% CI: $[+0.61, +3.78]~pp$). A paired t-test confirmed this improvement was statistically significant ($t(15) = 2.95,~p = 0.010$).
This overall pattern provides statistically significant evidence that decay is not universal in LLM-generated benchmarks, supporting our central thesis that temporal decay patterns are sensitive to benchmark formulation rather than solely reflecting intrinsic contamination of underlying source materials.

\paragraph{Error Mode Analysis.}
We conducted a manual inspection of 500 randomly sampled incorrect responses to reveal three major error types: misuse of theory or formulas (51\%), where models apply inappropriate theoretical principles; misinterpretation of problem statements (42\%), where models misunderstand requirements; and miscalculation (7\%), involving computational mistakes in multi-step derivations. Theory misuse errors often involve selecting superficially relevant but contextually inappropriate formulas, for instance, an LLM applying Fubini's Theorem without verifying the required integrability condition: $\iint |f(x,y)| \, dx \, dy < \infty$.

\begin{figure*}[t!]
    \centering
    \includegraphics[width=\linewidth]{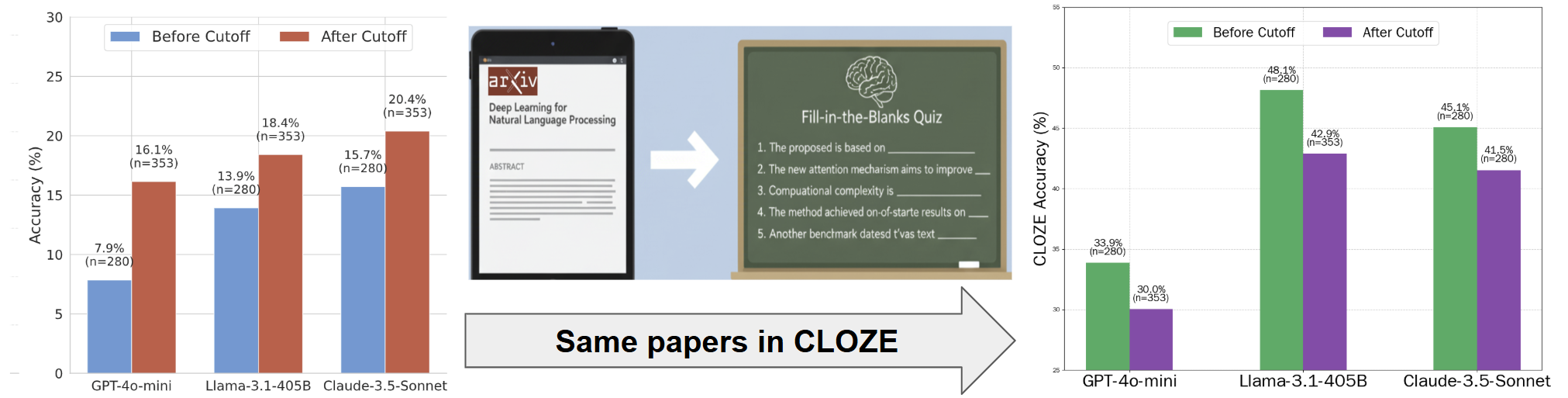}
    \caption{Temporal analysis on the same arXiv source material under two benchmark construction methods: LLM-synthesized multi-step questions (left) versus directly retrieved verbatim/cloze questions (right). Dashed vertical lines mark the evaluated models' respective knowledge cutoff dates.}
    \label{fig:realmath_comparison_full}
\end{figure*}

\section{Impact of benchmark construction on temporal pattern of model performance}
In Section 3, we have validated that the lack of post-cutoff performance decay is indeed a consistent pattern across various frontier model families.
We hypothesize that the main driver behind this phenomenon lies in the process of LLM generation from arXiv papers, which transformed the underlying material in such a way that evaluated models can no longer recognize them even if the source papers may have been included into the training corpora.

\subsection{Validation on cloze questions}
In this section, our aim is to test this hypothesis by constructing questions based on the very same arXiv papers without LLM generation.

\paragraph{Methodology.}
Specifically, we create 5 cloze (fill-in-the-blank) questions per paper by masking semantically meaningful terms from its abstract while preserving grammatical structure~(Sample questions in Table~\ref{tab:theorem_qa_conversion}). Each blank represents substantive content rather than easily inferred grammatical elements from context such as ``a/an''.
This process is performed by OpenAI-o4-mini with high reasoning effort and is validated manually.
For evaluation, we employ multiple metrics: BLEU~\cite{BLEU}, and ROUGE-L~\cite{ROUGE} scores and another LLM judge (with an independent o4-mini instance) with binary scoring (1 if answer matches the blanked phrase, 0 otherwise), cross-validated on 500 random instances with 94\% inter-annotator agreement to human annotation.

\paragraph{Results.}
We evaluated questions from 400 papers in our dataset on 3 models (o4-mini, Gemini-2.0-Flash, Llama-4-Scout) from our study (Table~\ref{tab:cloze_testoftime}), as well as questions from all the papers in RealMath~\cite{zhang2025realmath} on the models used in their work (Table~\ref{tab:cloze_realmath}).
 
In sharp contrast to the absence of decay in LLM-generated QA, we observe a predominant trend of visible post-cutoff performance decay on cloze questions across evaluated models and metrics.

\begin{table}[h!]
\centering
\setlength{\tabcolsep}{10pt}
\renewcommand{\arraystretch}{0.95}
\caption{Performance decay on CLOZE questions constructed from RealMath~\cite{zhang2025realmath} arXiv abstracts.}
\label{tab:cloze_realmath}
\resizebox{\columnwidth}{!}{
\begin{tabular}{lccc}
\hline
\textbf{Model} & \textbf{Pre (\%)} & \textbf{Post (\%)} & \textbf{Gap (pp)} \\
\hline
\multicolumn{4}{c}{\textbf{(a) LLM-Judge}} \\
\hline
GPT-4o-mini & 33.86 & 30.03 & -3.83 \\
Llama-3.1-405B & 48.14 & 42.89 & -5.25 \\
Claude-3.5-Sonnet & 45.07 & 41.53 & -3.54 \\
\hline
\multicolumn{4}{c}{\textbf{(b) ROUGE-L}} \\
\hline
GPT-4o-mini & 39.10 & 35.24 & -3.85 \\
Llama-3.1-405B & 49.42 & 44.21 & -5.22 \\
Claude-3.5-Sonnet & 48.75 & 43.76 & -4.98 \\
\hline
\multicolumn{4}{c}{\textbf{(c) BLEU}} \\
\hline
GPT-4o-mini & 16.43 & 14.41 & -2.02 \\
Llama-3.1-405B & 24.32 & 21.28 & -3.04 \\
Claude-3.5-Sonnet & 16.55 & 9.95 & -6.60 \\
\hline
\end{tabular}
}
\end{table}

\begin{table}[htbp]
\centering
\setlength{\tabcolsep}{10pt}
\renewcommand{\arraystretch}{0.95}
\caption{Performance on CLOZE questions synthesized in this work.}
\label{tab:cloze_testoftime}
\resizebox{\columnwidth}{!}{
\begin{tabular}{lccc}
\hline
\textbf{Model} & \textbf{Pre (\%)} & \textbf{Post (\%)} & \textbf{Gap (pp)} \\
\hline
\multicolumn{4}{c}{\textbf{(a) LLM-Judge}} \\
\hline
o4-mini & 49.70 & 49.00 & -0.70 \\
Gemini-2.0-Flash & 40.00 & 37.70 & -2.30 \\
Llama-4-Scout & 34.20 & 32.70 & -1.50 \\
\hline
\multicolumn{4}{c}{\textbf{(b) ROUGE-L}} \\
\hline
o4-mini & 47.93 & 47.68 & -0.25 \\
Gemini-2.0-Flash & 41.77 & 39.78 & -1.99 \\
Llama-4-Scout & 35.59 & 35.02 & -0.57 \\
\hline
\multicolumn{4}{c}{\textbf{(c) BLEU}} \\
\hline
o4-mini & 23.71 & 23.40 & -0.30 \\
Gemini-2.0-Flash & 17.96 & 17.68 & -0.28 \\
Llama-4-Scout & 15.33 & 15.84 & +0.51 \\
\hline
\end{tabular}
}
\end{table}

\subsection{Validation on LiveCodeBench}
To validate that LLM-driven transformation consistently acts as a confounder beyond arXiv papers, we examine LiveCodeBench~\cite{LiveCodeBench}, which has previously reported clear post-cutoff performance decay.

\paragraph{Methodology.}
We transform LiveCodeBench-Release-v1 (with data released between May 2023 and Mar 2024 containing 400 problems) using o4-mini with high reasoning effort to transform the problem statements while maintaining that the same algorithmic solutions apply to both transformed and original problems equally.
We refer to this transformed dataset as \textbf{PerturbLiveCodeBench} following the nomenclature of previous work in this field~\cite{MATH-Perturb}. Detailed perturbation prompts and sample questions before vs. after transformation are provided in Appendix~\ref{appendix:livecodebench_prompts}.

\begin{figure*}[t!]
    \centering
    \includegraphics[width=\linewidth]{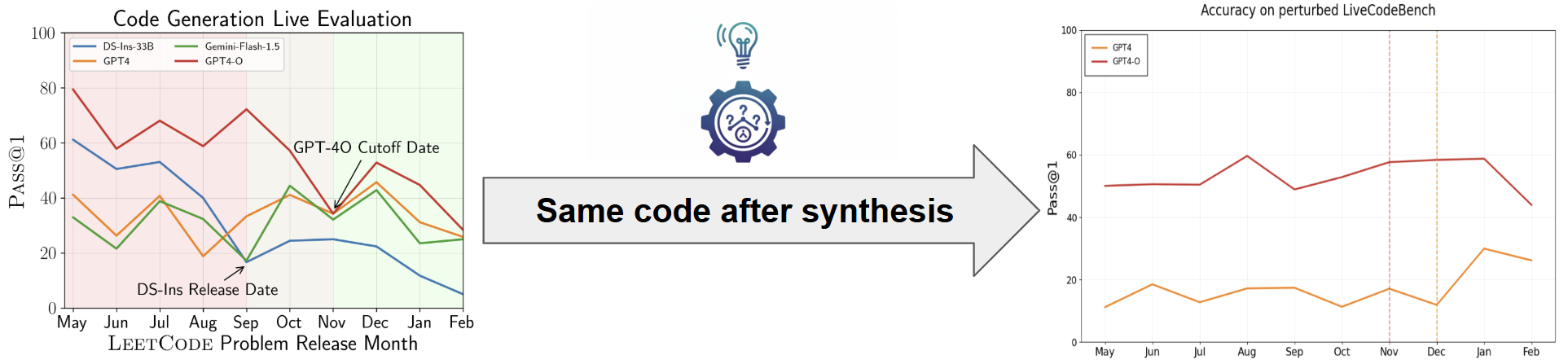}
    \caption{Validation Experiment on LiveCodeBench: temporal decay in the original LiveCodeBench~\cite{LiveCodeBench} (left) versus the same problems after OpenAI-o4-mini transformation (right), which preserves the underlying solution code. Dashed vertical lines mark the evaluated models' respective knowledge cutoff dates.}
    \label{fig:livecode_comparison_full}
\end{figure*}

\paragraph{Results.}
Following the original evaluation protocol of~\citet{LiveCodeBench}, we evaluate GPT-4o and GPT-4 on PerturbLiveCodeBench as they are the same models that previously reported post-cutoff performance decay.\footnote{Gemini-1.5 and DS-Ins-33B were deprecated by model developers at the time of experimentation.}
As shown in Figure~\ref{fig:livecode_comparison_full}, PerturbLiveCodeBench no longer exhibits a clear post-cutoff performance decay.
This stark contrast demonstrates that LLM-based question generation systematically removes post-cutoff performance decay even on problems that clearly exhibit decay in their original form.
Crucially, this validation experiment is performed in a completely different domain (coding) and benchmark construction paradigm (competitive programming), which further validates that this pattern is not unique to arXiv papers.

\subsection{Validation on Wiki-based QA}
We perform an additional validation experiment using a Wiki-based FreshQA-style~\cite{vu-etal-2024-freshllms} setting.

\paragraph{Methodology.}
Inspired by FreshQA-style setting, we crawl verbatim dated event text from \emph{Wikipedia Current Events archives} and construct dated multiple-choice questions from these event lines, spread uniformly across temporal windows of pre- and post-cutoff. We use o4-mini to apply semantic transformations only to the question statement while preserving the corresponding options and final answer. We evaluate GPT-3.5-turbo, GPT-4, and GPT-4o-mini on both the original and transformed variants.

\paragraph{Results.}
All 3 models exhibit significant post-cutoff performance decay on original dataset of the MCQs. GPT-3.5-turbo demonstrates $-2.65~pp$ performance gap; GPT-4 shows a gap of $-1.04~pp$; GPT-4o-mini exhibits the largest gap of $-7.59~pp$ on the original variants. We observe a stark contrast when evaluated on the LLM-transformed variants, where GPT-3.5-turbo shows a performance gap of $-0.62~pp$; GPT-4 has an astounding increase in post-cutoff performance of $+2.81~pp$; while GPT-4o-mini shows a change of $-4.99~pp$. This observation (Table~\ref{tab:wikiQAresults}) underscores the impact of benchmark construction method on temporal patterns, generally altering the pattern of post-cutoff performance decay significantly.

\begin{table}[htbp]
\centering
\caption{Comparison of LLM performance on original versus transformed variants of Wiki-based QA. Post-cutoff performance (\emph{Post $-$ Pre}) shows significant change.}
\label{tab:wikiQAresults}
\small
\begin{tabular}{lcc} 
\hline
\textbf{Model} & \textbf{Original (pp)} & \textbf{Transformed (pp)} \\
\hline
GPT-3.5-turbo & -2.65 & -0.62 \\
GPT-4 & -1.04 & +2.81 \\
GPT-4o-mini & -7.59 & -4.99 \\
\hline
\end{tabular}
\end{table}

\section{Mechanistic Interpretability Analysis}
\begin{figure}[t]
\centering
\includegraphics[width=\linewidth]{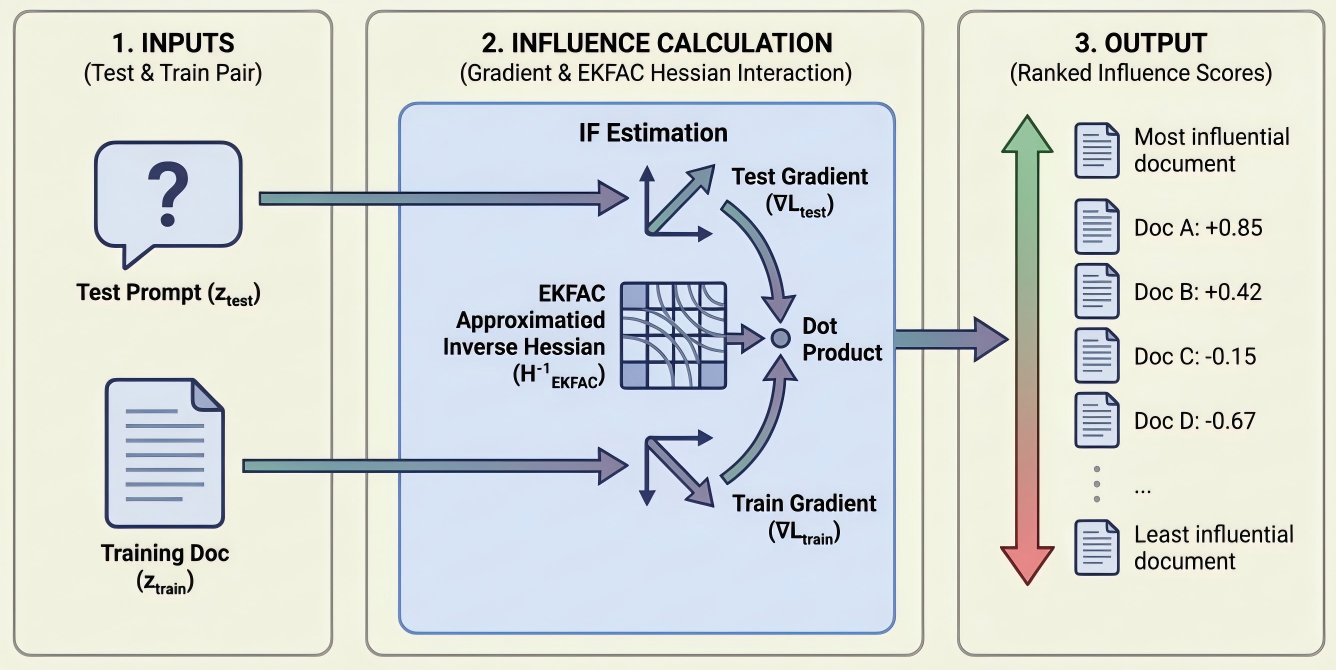}
\caption{Overview of our influence function analysis framework.}
\label{fig:influence}
\end{figure}

In the previous sections, we focus on temporal analysis on the level of model performance by comparing accuracy score in pre- vs. post-cutoff questions.
In this section, our aim is to study the underlying mechanism behind this phenomenon by studying the internal working of LLMs using mechanistic interpretability methods.
Influence functions~\cite{koh2020understandingblackboxpredictionsinfluence,grosse2023studyinglargelanguagemodel} offer a powerful tool to trace and rank the most responsible data points for a given model input-output pair as shown in Figure~\ref{fig:influence}.
Specifically, our aim is to show that based on the same contaminated paper, LLM-transformed questions are much harder for models to trace back to their source paper than cloze questions.

\subsection{Review of Influence Functions}
Let $\mathcal{D}$ denote the training dataset for model $\mathcal{M}$ parameterized by $\theta$. The influence function quantifies how a training document $z \in \mathcal{D}$ impacts the model's prediction on a test input-output pair $z_{\text{test}}=(z_p,z_c)$, where $z_p$ is the model \emph{input} (prompt/prefix tokens) and $z_c$ is the model \emph{output} (target completion tokens).
Following the standard formulation~\cite{koh2020understandingblackboxpredictionsinfluence}, we approximate the change in loss at $z_{\text{test}}$ if training point $z$ were up-weighted by an infinitesimal $\epsilon$ as
\begin{equation}
\mathcal{I}(z, z_{\text{test}})
~=~-
\nabla_{\theta} L(z_{\text{test}}, \hat{\theta})^\top
H_{\hat{\theta}}^{-1}
\nabla_{\theta} L(z, \hat{\theta}),
\end{equation}
where $L$ is the loss function, $\hat{\theta}$ is the empirical risk minimizer, and $H_{\hat{\theta}}$ is the Hessian of the loss. Due to the high dimensionality of $\theta$ in LLMs, exact inverse-Hessian computation is intractable. In our implementation, we follow Kronfluence~\cite{grosse2023studyinglargelanguagemodel} and define $f(\theta)=\log p(z_c\mid z_p;\theta)$, the conditional log-likelihood of the output tokens $z_c$ given the input tokens $z_p$. The influence score is then approximated as
\begin{equation}
I_f(z) \;\approx\; - \nabla_\theta f(\theta^s)^\top \, (G + \lambda I)^{-1} \, \nabla_\theta L(z, \theta^s),
\end{equation}
where $\theta^s$ are the pretrained weights, $G$ is a (block-diagonal) Gauss--Newton or Fisher curvature matrix, and $\lambda>0$ is a damping constant for numerical stability. We compute the required inverse-curvature vector products using EK-FAC~\cite{george2021fastapproximatenaturalgradient} as implemented in Kronfluence. Additional details are provided in Appendix~\ref{appendix:influence_details}.

\subsection{Experiments}
Most of the mainstream open-weight models (such as DeepSeek, Llama, Qwen) release model weights but not their training data, which makes it impossible to provably tell whether a given document is part of its training data.
Therefore, we make use of OLMo2-7B-Instruct~\cite{olmo20252olmo2furious}, one of the best-performing open-data LLMs with publicly available training dataset.
We select 40 arXiv papers from the training corpora of OLMo2-7B-Instruct\footnote{This sample size represents the maximum processable on a single GH200 GPU within 24 hours. We use instruct model, as the base version has insufficient instruction-following capability to properly answer our questions in this setting.}, which are provably contaminated since they form part of the training documents.

For each paper, we create both a cloze question and an LLM-generated question based on the same framework used in our previous experiments.
From the full training corpus, we randomly sample 10,000 documents that includes the 40 arXiv papers we used for evaluation and rank the top 100 most influential training documents for each query. Implementation details are in Appendix~\ref{appendix:influence_details}.

\begin{table}[h!]
\centering
\caption{Top-$K$ hit rates where influence function methods successfully identified the contaminated documents among the Top-$K$ most influential training documents. The sample size for this metric is 40 papers.}
\label{tab:influence_results}
\begin{tabular}{lcc} 
\hline
\textbf{Hit Rate} & \textbf{cloze} & \textbf{LLM-generated QA} \\
\hline
Top-1 & 77.5\% & 17.5\% \\
Top-3 & 100.0\% & 25.0\% \\
\hline
\end{tabular}
\end{table}

\subsection{Results}
We define \textbf{Top-K Hit Rate} as the percentage of queries for which the corresponding source paper appears among the $K$ most influential documents. For example, Top-1 Hit Rate indicates the percentage where the source paper is identified as the single most influential document.

We report the hit rate for Top-1 and Top-3 in Table~\ref{tab:influence_results}.
For cloze questions, the evaluated model was able to identify the corresponding source document as the single most influential training data point with a top-1 hit rate of 77.5\%.
Furthermore, the correct source document could be 100\% identified within Top-3 most influential training documents, which indicated clear recall from training corpora when trying to answer these questions.

On the other hand, LLM-generated QA based on the very same papers in the training corpora yielded a significantly lower rate with Top-1 hit rate at only 17.5\% and Top-3 hit rate at 25.0\%.
It's worth noting that the sample size for this result is very limited due to the compute-intensive nature of influence function analysis, a meaningful direction for future investigations could be scaling up this type of analysis on larger datasets to obtain more statistically significant results.

\section{Discussion}
Overall, we provide a complementary analysis at both the model-performance level and the level of internal mechanisms:

At the performance level, we first validated the lack of post-cutoff decay across 8 frontier models over 26 months using LLM-generated questions based on arXiv papers. But when we convert these same papers to cloze questions, post-cutoff decay patterns emerge on the same models to show that the key factor lies in the LLM generation process itself rather than the underlying source material.

We further validated this hypothesis beyond arXiv papers using LiveCodeBench~\cite{LiveCodeBench} (which has reported clear post-cutoff performance decay) and FreshQA-type~\cite{vu-etal-2024-freshllms} questions based on Wikipedia. We show that simply rephrasing these questions with LLM without changing their underlying solutions could also effectively remove post-cutoff performance decay.

At the mechanistic level, influence function analysis further confirmed our hypothesis by showing that models can identify the responsible source documents as the most influential when given cloze questions, but much less so when given LLM-generated questions.

Our key thesis centers around the fragility of post-cutoff performance decay as a temporal contamination signal. We specifically showed that simple LLM generation/transformation could effectively remove post-cutoff performance decay 
from questions that previously reported such decay~\cite{LiveCodeBench}.
Compared to previous work that explicitly adds new knowledge after knowledge cutoff~\cite{AntiLeakBench}, we further demonstrated that even for the same underlying material, transformation alone could already effectively impact the temporal pattern of model performance without incorporating any new additional information.

\section{Conclusion and Outlook}
We call on the community to carefully consider the reliability of post-cutoff performance decay as a temporal signal for benchmark contamination, as such signal is highly sensitive to benchmark construction methodology, which could yield very different temporal pattern even based on the same source material.

It's worth noting that we do \textbf{not} claim that absence of performance decay means absence of contamination. Indeed, our influence function analysis reveals that LLM-generated questions from provably contaminated papers show a clear lack of post-cutoff decay. Our counter-intuitive finding affirms that post-cutoff performance decay as a temporal signal for contamination is highly sensitive to how the questions are constructed.
To elucidate the practical implication of this work, we offer a few concrete recommendations for future work:
\begin{itemize}
  \item Document detailed methodology for benchmark construction and carefully examine how the different construction methods may impact evaluation results;
  \item Recognize that the temporal pattern of model performance before vs. after cutoff is a highly sensitive probe that should be validated against other contamination detection methods;
  \item Develop more robust contamination detection methods that report model performance as more than a scalar score, while also factoring in the level of fluctuation when the same problem is transformed without changing its core.
\end{itemize}

In conclusion, we hope this work could serve as a thought-provoking piece to motivate future work on more robust contamination detection methods and more reliable benchmarking practices.

\section*{Limitations}
First, our temporal analysis is based on a 26-month time window with a limited amount of papers due to budgetary concerns (processing long arXiv papers will incur considerable token usage costs).
Our framework remains scalable to any arbitrary time window, where the only bottleneck factor is the number of papers available on arXiv in any user-defined time window.
Similarly, influence function analysis is highly compute-intensive in nature, which is why we chose to limit our scope of experimentation to 40 papers.
Lastly, we also acknowledge that the counter-intuitive rise of model performance after cutoff dates in some models remains a puzzling open question for future investigations, one potential explanation could be that the exponentially increasing number of papers on arXiv in recent years may contain much more trivial ones, subsequently making the questions considerably easier to answer.
We also recognize that the variance of arXiv papers and thus the difficulty of our synthesized questions may be a confounder in our analysis, which we strive to mitigate using a large sample size and relatively long time window in the temporal analysis.


\section*{Acknowledgements}
This manuscript is based on work supported in part by Schmidt Sciences; by the German Federal Ministry of Education and Research (BMBF) and T\"ubingen AI Center (FKZ: 01IS18039B), by Machine Learning Cluster of Excellence (EXC 2064/1, Project 390727645); and by Coefficient Giving.

\newpage
\appendix
\cleardoublepage
\newcommand{\appendixplotheight}{0.38\textheight}
\newcommand{\appendixsubplotheight}{0.30\textheight}
\section{Responsible NLP Statements}
\paragraph{Risk and AI Usage Statement}
Currently we don't see any ethical risk of this study. Our dataset contains no personal identifying information or offensive content and we use datasets which are all publicly accessible under the CC-BY-4.0 license for non-commercial academic research.
AI assistants (such as LLMs) were used as aids for paper writing in grammatical checks and help with visualization such as creating flaticons in figures. We used PaperMentor~\cite{liu-etal-2026-papermentor} to get feedback on our structure of the paper and overall writing.

\paragraph{Experimental Data Statistics}
Our evaluation subset contains 1,643 QA pairs (856 mathematics, 787 physics) synthesized from 20,277 arXiv papers (May 2023--June 2025). Initial corpus (4,235 mathematics, 16,042 physics) was filtered to 1,098 papers (579 mathematics, 519 physics).

\section{Influence Function Analysis Details}
\label{appendix:influence_details}

This section provides the technical details for the influence function analysis presented in the main paper. We use influence functions~\citep{grosse2023studyinglargelanguagemodel,ruis2025proceduralknowledgepretrainingdrives} to mechanistically investigate the relationship between data points in pretraining documents and model output on different input queries (in this case the difference of CLOZE vs. synthesized QA questions).

\subsection{Methodology}
For each arXiv paper in our dataset, we compute influence scores for two paired questions: (1) a retrieval-based CLOZE question testing direct recall from the abstract, and (2) a synthesis-based question by reasoning models requiring multi-step problem solving.

\subsection{Preliminaries}
\label{preliminaries}

We briefly review influence functions and the EK-FAC (Eigenvalue-corrected Kronecker-Factored Approximate Curvature) approximations before presenting our findings. Influence functions estimate how individual training points affect model predictions, and EK-FAC provides a scalable second-order approximation that makes this analysis feasible for large LLMs.

\paragraph{Influence functions.} 
Influence functions estimate how upweighting or removing a training example affects model predictions \citep{koh2020understandingblackboxpredictionsinfluence}. 
For a query $z_{\text{test}} = (z_p, z_c)$, we are interested in the influence of a candidate training sequence $z \in \mathcal{D}$ on the model's conditional log-likelihood.
Here, $z_p$ denotes the \emph{prefix} (the model \emph{input}/prompt tokens), and $z_c$ denotes the \emph{completion} (the model \emph{output}/target continuation tokens) whose likelihood is evaluated under the model.
We define
$f(\theta) = \log p(z_c \mid z_p; \theta)$. The influence is approximated as
\begin{equation}
I_f(z) \;\approx\; - \nabla_\theta f(\theta^s)^\top \, (G + \lambda I)^{-1} \nabla_\theta L(z, \theta^s),
\end{equation}
where $\theta^s$ are the pretrained model weights, $L$ is the training loss, $G$ is the Gauss-Newton Hessian, 
and $\lambda > 0$ is a damping constant for numerical stability.  

\paragraph{Method used: EK-FAC approximation.} 
We use the \emph{Eigenvalue-Corrected Kronecker-Factored Approximate Curvature (EK-FAC)} method \citep{george2021fastapproximatenaturalgradient} integrated within the Kronfluence architecture \cite{grosse2023studyinglargelanguagemodel} to perform influence function analysis. EK-FAC leverages the Kronecker structure of the Fisher information matrix (or Gauss-Newton Hessian) in deep networks, 
enabling efficient eigendecomposition and inversion. Specifically:
\begin{itemize}
    \item The Kronecker factors $A$ and $S$ (capturing input and output covariances of each layer) admit tractable eigendecompositions.  
    \item Their Kronecker product $A \otimes S$ is diagonalized via the eigenvectors of $A$ and $S$, yielding a diagonal approximation $\Lambda$ whose entries capture variance in the projected pseudo-gradients.  
    \item Damping is naturally incorporated by adding $\lambda$ to the eigenvalues, so IHVPs reduce to rescaling in this eigenbasis.  
\end{itemize}

Let $Q_{AS} = Q_A \otimes Q_S$ denote the Kronecker eigenbasis.

Formally, EK-FAC approximates $G$ as
\begin{equation}
G \;\approx\; Q_{AS} \, \Lambda \, Q_{AS}^\top,
\end{equation}
and the damped IHVP as
\begin{equation}
(G + \lambda I)^{-1} v \;\approx\; Q_{AS} \, (\Lambda + \lambda I)^{-1} \, Q_{AS}^\top v
\end{equation}

Once the eigendecomposition is computed, IHVPs can be applied efficiently for many queries, making EK-FAC well-suited to influence-function analysis in large-scale transformer LMs.

\subsection{Experimental Setup}
\paragraph{Model and Corpus}
We use the OLMo2-7B-Instruct model from the OLMo 2 collection for influence function analysis. The pretraining corpus consists of diverse documents spanning scientific publications, web content, and technical documentation. For computational tractability, we sample 10,000 documents evenly distributed across the corpus to approximate the Hessian matrix.

\paragraph{Computational Justification}
The Hessian matrix encodes second-order optimization information across all model parameters. While computing the exact Hessian would require processing the entire training corpus, the primary computational bottleneck lies in gradient calculations for each document. Prior work~\citep{ruis2025proceduralknowledgepretrainingdrives} (Appendix A.2) demonstrates that influence scores remain highly correlated even when computed using a subset of documents for Hessian approximation. Our 10,000-document sample provides a tractable yet representative estimate without materially affecting the validity of our findings, while also minimizing computational resource usage for environmental considerations.

\subsection{Results Interpretation}
\paragraph{Influence Score Distribution}
Influence scores quantify how much each pretraining document affects the model's output for a given query. In our experiments, scores range from approximately 25-27 million for the most influential documents down to 8-16 million for the least influential among the top 100. This variation reflects differing document impact: higher scores indicate that removing or modifying the document would induce larger changes in model behavior, while lower scores correspond to comparatively smaller influence.

\section{Details for Validation on LiveCodeBench}\label{appendix:livecodebench_prompts}
For the perturbed LiveCodeBench~\cite{LiveCodeBench} experiment, we use o4-mini to generate semantically equivalent variations of the original coding problems while preserving algorithmic complexity, enforcing (i) \emph{algorithmic equivalence} (the same algorithmic approach and computational complexity as the original), (ii) \emph{consistent transformation} throughout problem statements and test cases (e.g., \texttt{abc} $\rightarrow$ \texttt{XYZ}, \texttt{acb} $\rightarrow$ \texttt{XZY}), and (iii) \emph{test case validity}, where expected outputs are updated to reflect the input transformations; this perturbation approach enables controlled experiments distinguishing genuine problem-solving abilities from memorization of specific problem formulations.
We detail the prompts below: 

\begin{lstlisting}[basicstyle=\scriptsize\ttfamily]
You are tasked with creating a variation of the following programming problem.

The variation should:
1. Keep the exact same algorithmic approach and complexity
2. Change variable names, function names, and context
   (e.g., if it uses 'abc', use something like 'XYZ')
3. Modify specific values in test cases consistently with the context change
4. Maintain the same difficulty level and logic

Original Problem:
{problem_text}

Original Test Examples:
{test_examples}

Provide the perturbed problem AND perturbed test cases in the following JSON format:
{
    "problem_statement": "...",
    "test_cases": [
        {"input": "...", "output": "...", "testtype": "stdin"}
    ]
}

Make sure to perturb ALL test values consistently. If the original uses 'abc', 'acb', 'bac',
etc., and you change to 'XYZ', then use 'XYZ', 'XZY', 'YXZ', etc. correspondingly.
\end{lstlisting}

\subsection{PerturbLiveCodeBench Examples}\label{appendix:livecodebench_examples}
We demonstrate representative examples of original problems and their transformed variants generated using the perturbation prompt described in Section~\ref{appendix:livecodebench_prompts}. The goal of these transformations is to preserve the underlying algorithmic structure and solution strategy while altering surface-level characteristics such as variable names, semantic framing, and symbolic representations.

\begin{table}[h]
\centering
\small
\caption{Examples of original vs. transformed problems from  LiveCodeBench.}
\label{tab:livecodebench_examples}

\renewcommand{\arraystretch}{1.2}

\begin{tabular}{p{0.95\linewidth}}
\toprule

\textbf{Example A} \\[0.3em]

\textbf{Original:} \\[0.2em]
You need to find two numbers in an array that add up to a target sum. 
Return the indices of the two numbers. \\[0.4em]

\textbf{Transformed:} \\[0.2em]
You are given a list of integer weights. Identify two weights whose 
combined total matches a specified target value. Return the positions 
of the two weights. \\[0.6em]

\midrule

\textbf{Example B} \\[0.3em]

\textbf{Original:} \\[0.2em]
Given two sorted arrays nums1 and nums2, find the median of the two 
sorted arrays. \\[0.4em]

\textbf{Transformed:} \\[0.2em]
You are given two sorted sequences dataX and dataY. Determine the median 
of the merged sorted collection. \\[0.6em]

\midrule

\textbf{Example C} \\[0.3em]

\textbf{Original:} \\[0.2em]
Given a string containing only '(' and ')', find the length of the 
longest valid parentheses substring. \\[0.4em]

\textbf{Transformed:} \\[0.2em]
Given a sequence containing only '[' and ']', find the length of the 
longest valid bracket substring. \\

\bottomrule
\end{tabular}
\end{table}

These examples illustrate three common categories of perturbations: (i) lexical substitution (e.g., \texttt{array} $\rightarrow$ \texttt{list}, \texttt{numbers} $\rightarrow$ \texttt{weights}), (ii) variable renaming, and (iii) symbol substitution (e.g., parentheses replaced with brackets). Despite these changes, the core computational tasks correspond to well-known problems such as two-sum, median of two sorted arrays, and longest valid parentheses. 

Therefore, we emphasize that these transformations are intentionally lightweight and do not introduce additional reasoning complexity. As such, they provide a controlled setting for our validation experiment without conflating it with task difficulty.

\section{Additional Visualizations}\label{appendix:additional_plots}

This section provides supplementary visualizations that contextualize the temporal analysis results in the main paper. We include (i) pre/post-cutoff averages, (ii) full month-by-month trajectories, and (iii) aggregated $n$-month windows to reduce variance from uneven monthly sample sizes.

\subsection{Mean Accuracy Trends}\label{appendix:math_results}
The summary statistics below report normalized accuracy averaged over all pre-cutoff months versus all post-cutoff months for each model (with the sign convention $\text{Gap} = \text{Post} - \text{Pre}$). The accompanying plots visualize the same comparison, highlighting that the observed differences are small and not systematically negative after the cutoff.

\begin{figure*}[h!]
    \centering
    \begin{subfigure}[htbp]{0.9\linewidth}
        \centering
        \includegraphics[width=\linewidth,height=\appendixsubplotheight,keepaspectratio]{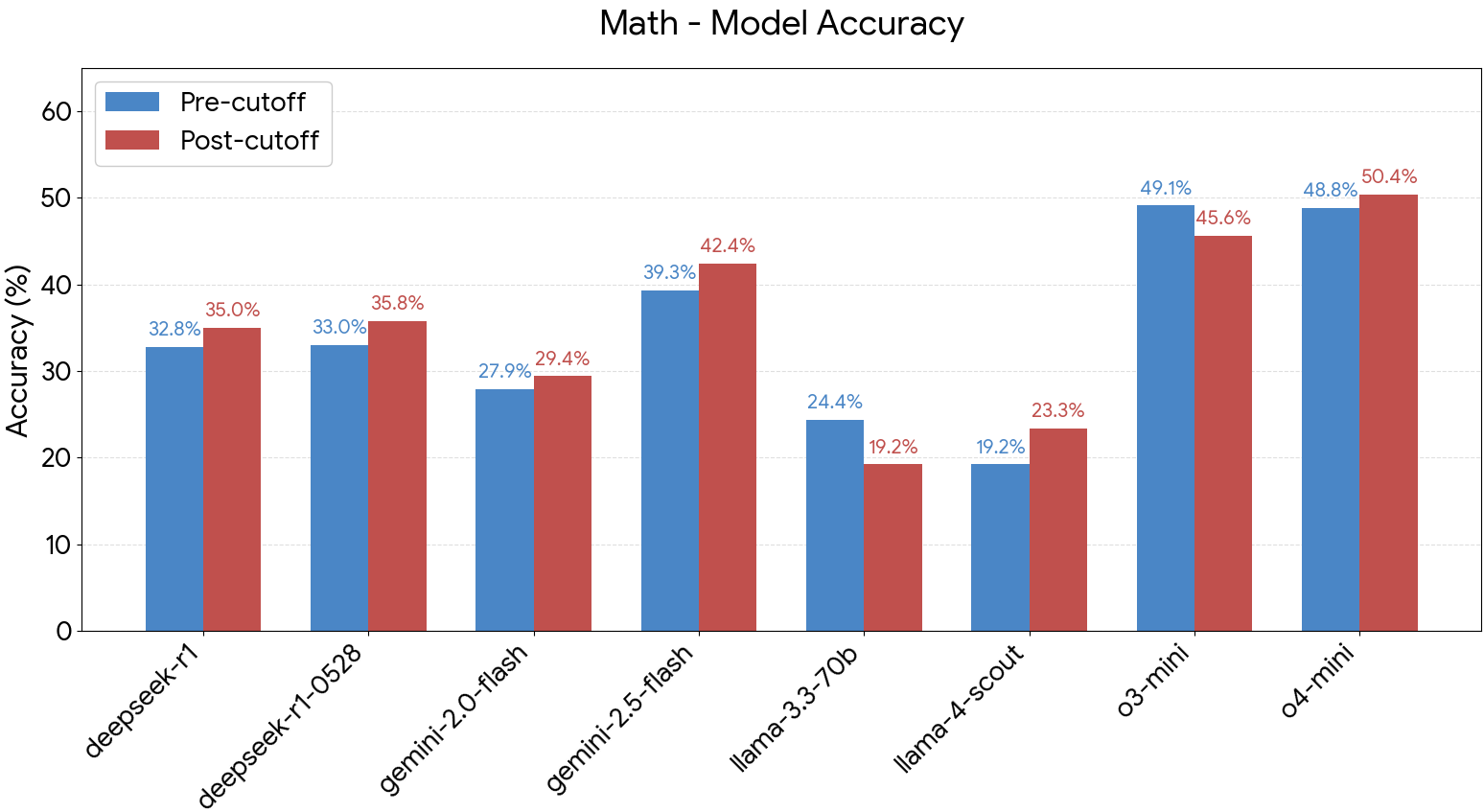} 
        \label{fig:math_mean}
    \end{subfigure}
    \hfill
    \begin{subfigure}[htbp]{0.9\linewidth}
        \centering
        \includegraphics[width=\linewidth,height=\appendixsubplotheight,keepaspectratio]{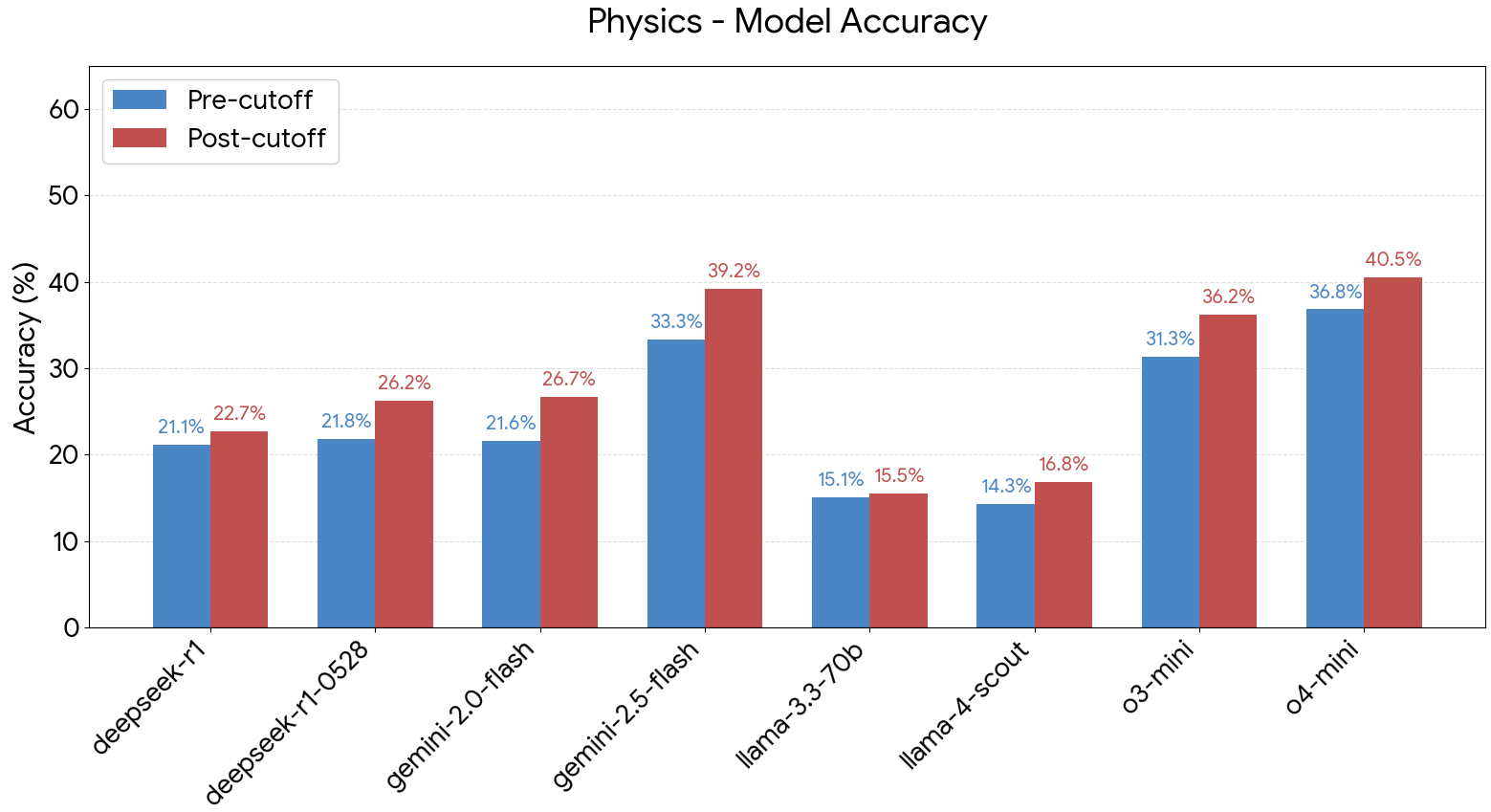} 
        \label{fig:phys_mean}
    \end{subfigure}
    \caption{Mean accuracy trends before (blue) versus after (red) knowledge cutoff dates for the Mathematics and Physics domain.}
    \label{fig:math_phys_mean}
\end{figure*}

\begin{table*}[htbp]
    \centering
    \caption{Pre- versus post-cutoff accuracy in the Math domain. Difference $(Gap~(pp))$ is reported as increase in post-cutoff accuracy as compared to pre-cutoff accuracy.}
    \label{tab:pre_post_math}
    \begin{tabular}{lccc}
    \hline
    \textbf{Model} & \textbf{Pre-cutoff (\%)} & \textbf{Post-cutoff (\%)} & \textbf{Gap (pp)} \\
    \hline
    Deepseek-R1        & 32.8\% & 35.0\% & +2.2 \\
    Deepseek-R1-0528   & 33.0\% & 35.8\% & +2.8 \\
    Gemini-2.0-Flash   & 27.9\% & 29.4\% & +1.5 \\
    Gemini-2.5-Flash   & 39.3\% & 42.4\% & +3.1 \\
    Llama-3.3-70B      & 24.4\% & 19.2\% & -5.2 \\
    Llama-4-Scout      & 19.2\% & 23.3\% & +4.1 \\
    o3-mini            & 49.1\% & 45.6\% & -3.5 \\
    o4-mini            & 48.8\% & 50.4\% & +1.6 \\
    \hline
    \end{tabular}
\end{table*}
Table~\ref{tab:pre_post_math} offers a compact view of these averages in the mathematics domain. Figure~\ref{fig:math_phys_mean} complements it by showing the mean pre- versus post-cutoff accuracy for both domains.

\subsection{Monthly Performance Trends}\label{appendix:monthly}
While the pre/post averages collapse time into two bins, the following plot shows the full month-by-month trajectories from May 2023 to June 2025. Dashed vertical lines indicate each model's reported cutoff month; visually, performance does not exhibit a consistent downward shift immediately after these boundaries.

\begin{figure*}[h!]
    \centering
    \includegraphics[width=\linewidth,keepaspectratio]{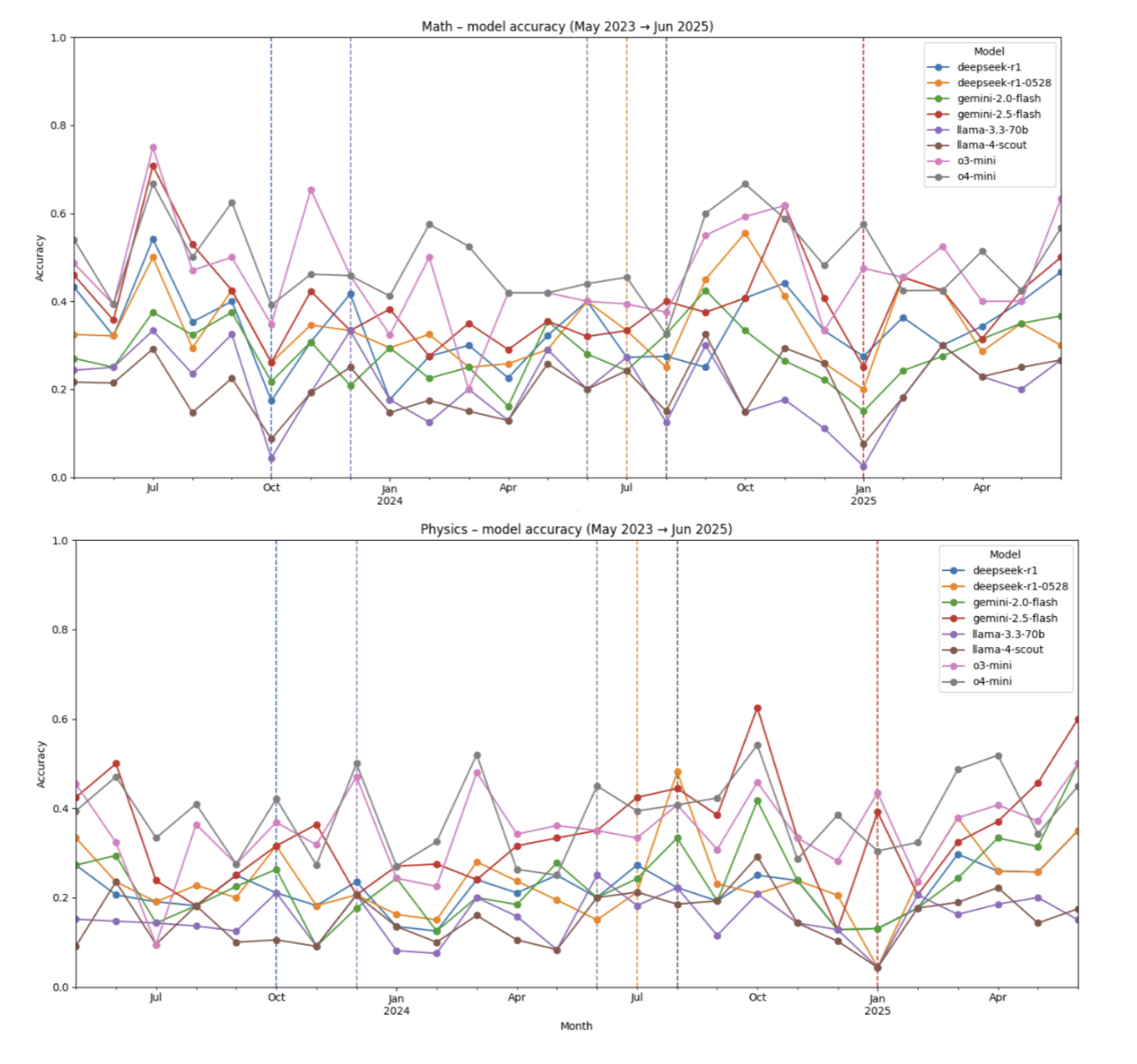}
    \caption{Model performance on synthesis-based questions by reasoning models across 26 months (May 2023 to June 2025) for mathematics (top) and physics (bottom) domains, with knowledge cutoff dates marked by dashed lines. Performance remains stable across all models' cutoff boundaries, with some models showing slight improvements post-cutoff.}
    \label{fig:monthly_performance}
\end{figure*}

\subsection{n-Month Aggregated Results}\label{appendix:aggregate_results}
To further mitigate month-level noise, we aggregate accuracy over fixed-width windows around the cutoff. For each model and each $n \in \{2,3,4,5\}$, we compute accuracy on the $n$ months immediately before the cutoff (denoted $nB$) and the $n$ months immediately after the cutoff (denoted $nA$). This view tests whether conclusions are sensitive to the exact choice of temporal window.

\begin{figure*}[h!]
    \centering
    \begin{subfigure}[htbp]{0.9\linewidth}
        \centering
        \includegraphics[width=\linewidth,height=\appendixsubplotheight,keepaspectratio]{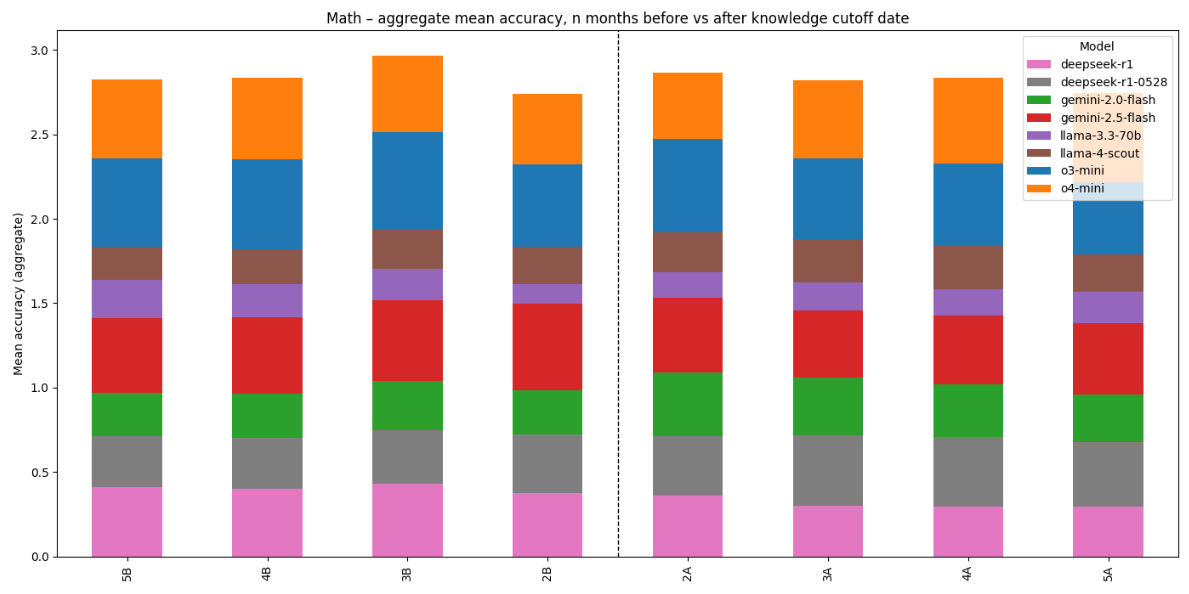} 
        \caption{Mathematics: Aggregated time windows}
        \label{fig:nmonth_math_agg}
    \end{subfigure}
    \hfill
    \begin{subfigure}[htbp]{0.9\linewidth}
        \centering
        \includegraphics[width=\linewidth,height=\appendixsubplotheight,keepaspectratio]{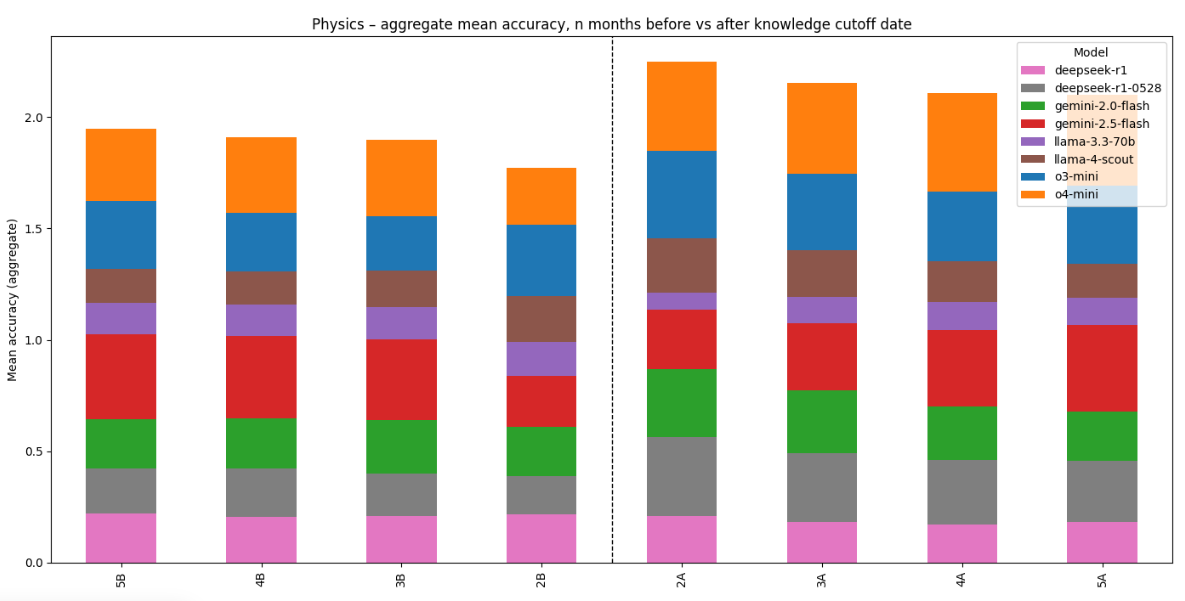} 
        \caption{Physics: Aggregated time windows}
        \label{fig:nmonth_phys_agg}
    \end{subfigure}
    \caption{Mathematics and Physics aggregated performance across multiple time windows (nB marks $n$ months before, nA marks $n$ months after cutoff). Each color represents a different model in question whereas each bar represents one time window for aggregation.}
    \label{fig:nmonth_math_phys_agg}
\end{figure*}

Overall, the aggregated windows lead to the same qualitative conclusion as the monthly plots: across models and domains, post-cutoff performance does not consistently drop relative to pre-cutoff performance, and in several cases it slightly increases.
\end{document}